# Privacy-Preserving Customer Support: A Framework for Secure and Scalable Interactions

Anant P. Awasthi[1], Girdhar G. Agarwal[5], Chandraketu Singh[2], Rakshit Varma[3], Sanchit Sharma[4]

[1, 3, 4] Optum Global Solutions, Noida, Gautam Buddha Nagar, Uttar Pradesh, India

[2] Jaipuria Institute of Management, Lucknow, Uttar Pradesh, India

[5] Ex-Professor, University of Lucknow, Lucknow, Uttar Pradesh, India

**Correspondence**: Anant Prakash Awasthi, Lead Data Scientist, Optum Global Solutions, Noida, Gautam Buddha Nagar, Uttar Pradesh, India; E-mail: anant.awasthi@outlook.com

**Abstract**:

The growing reliance on artificial intelligence (AI) in customer support has significantly improved operational efficiency and user experience. However, traditional machine learning (ML) approaches, which require extensive local training on sensitive datasets, pose substantial privacy risks and compliance challenges with regulations like the General Data Protection Regulation (GDPR) and California Consumer Privacy Act (CCPA). Existing privacy-preserving techniques, such as anonymization, differential privacy, and federated learning, address some concerns but face limitations in utility, scalability, and complexity.

This paper introduces the **Privacy-Preserving Zero-Shot Learning (PP-ZSL) framework**, a novel approach leveraging large language models (LLMs) in a zero-shot learning mode. Unlike conventional ML methods, PP-ZSL eliminates the need for local training on sensitive data by utilizing pre-trained LLMs to generate responses directly. The framework incorporates real-time data anonymization to redact or mask sensitive information, retrieval-augmented generation (RAG) for domain-specific query resolution, and robust post-processing to ensure compliance with regulatory standards. This combination reduces privacy risks, simplifies compliance, and enhances scalability and operational efficiency.

Empirical analysis demonstrates that the PP-ZSL framework provides accurate, privacy-compliant responses while significantly lowering the costs and complexities of deploying AI-driven customer support systems. The study highlights potential applications across industries, including financial services, healthcare, e-commerce, legal support, telecommunications, and government services. By addressing the dual challenges of privacy and performance, this framework establishes a foundation for secure, efficient, and regulatory-compliant AI applications in customer interactions.

**Keywords:**

Privacy-Preserving Zero-Shot Learning (PP-ZSL), Large Language Models (LLMs), Data Privacy and Anonymization, Zero-Shot Learning (ZSL), Regulatory Compliance (GDPR, CCPA), Retrieval-Augmented Generation (RAG)

# 1. Introduction

Artificial intelligence (AI) has transformed customer support by enabling real-time, intelligent assistance. However, these advancements come with challenges, particularly in ensuring data privacy. Traditional machine learning (ML) models require extensive local training on organization-specific datasets, often containing sensitive information like personally identifiable information (PII), financial data, or contractual details. This dependency increases privacy risks and complicates compliance with stringent regulations such as the General Data Protection Regulation (GDPR) and the California Consumer Privacy Act (CCPA).

Local training introduces vulnerabilities in data handling, storage, and sharing, creating opportunities for data breaches and misuse. Existing privacy-preserving methods, such as anonymization, differential privacy, and federated learning, offer partial solutions but face significant challenges. For instance, anonymization often reduces data utility, differential privacy involves trade-offs with model performance, and federated learning adds operational complexity and scalability concerns. As customer demands for both privacy and efficiency grow, organizations require innovative approaches that minimize data risks without compromising service quality.

This paper proposes a transformative solution by leveraging **Zero-Shot Learning (ZSL)** with large language models (LLMs). Unlike traditional ML methods that require fine-tuning or local training, ZSL relies on pre-trained LLMs to generate responses without additional data training. This approach eliminates the need to store or process sensitive data locally, inherently reducing privacy risks. By leveraging the generalization capabilities of LLMs, ZSL offers organizations a scalable and cost-effective solution for customer support that ensures both accuracy and data security.

To operationalize this concept, we introduce a **Privacy-Preserving Zero-Shot Learning (PP-ZSL)** framework. The framework features:

1. **Real-Time Data Anonymization**: Sensitive details like PII and financial information are redacted or masked before queries are sent to the LLM.
2. **Retrieval-Augmented Generation (RAG)**: For domain-specific queries, the model fetches information from secure, non-sensitive knowledge repositories.
3. **Post-Processing and Validation**: Responses are audited to ensure compliance with privacy policies and regulatory standards.

PP-ZSL not only minimizes the risks associated with traditional ML approaches but also simplifies compliance with regulations like GDPR and CCPA. By avoiding local data training, organizations can streamline audits and ensure adherence to privacy requirements, including the "right to be forgotten" and data minimization principles. The framework also supports operational goals, addressing challenges like latency, accuracy, and scalability in customer support.

The present work explores the limitations of current methods, details the PP-ZSL framework, and evaluates its efficacy through empirical case studies. The findings demonstrate how organizations can achieve a balance between cutting-edge AI performance and robust privacy protection. In doing so, PP-ZSL establishes a foundation for ethical, secure, and efficient AI-driven customer support systems.

## 2. Review of Literature

The intersection of machine learning, privacy preservation, and customer support has been extensively studied across various domains. This review synthesizes key contributions in privacy-preserving techniques, the evolution of large language models (LLMs), and their application in customer service, highlighting the gaps addressed by this publication.

### 2.1 Privacy Challenges in Machine Learning

Machine learning models require large volumes of data for training, raising concerns about privacy risks when using sensitive data, such as personally identifiable information (PII) or financial details. Traditional anonymization techniques, such as pseudonymization, often fail to ensure robust privacy due to the potential for re-identification (Rocher et al., 2019). Differential privacy, as introduced by Abadi et al. (2016), offers theoretical guarantees against data reconstruction but often comes at the cost of model utility. Federated learning (Kairouz et al., 2021) mitigates some privacy concerns by distributing model training to local devices; however, it introduces scalability and communication overhead challenges.

### 2.2 Regulatory Landscape and Privacy Compliance

Regulations such as the General Data Protection Regulation (GDPR) and the California Consumer Privacy Act (CCPA) emphasize the need for data minimization, the right to be forgotten, and accountability in automated decision-making (European Union, 2016; CCPA, 2018). These frameworks create a legal imperative for businesses to implement privacy-preserving practices, especially in customer support operations. Existing solutions often fall short in achieving compliance while maintaining high-quality service delivery.

### 2.3 Large Language Models and Zero-Shot Learning

Recent advancements in LLM (Large Language Models), such as GPT-3 and T5, have demonstrated the ability to perform complex tasks without task-specific training, a concept known as zero-shot learning (Brown et al., 2020; Raffel et al., 2020). These models leverage extensive pre-training on diverse datasets, enabling them to generalize across domains. Zero-shot learning reduces the dependency on local training data, presenting a promising avenue for privacy preservation in applications involving sensitive information. However, there are challenges in applying these models to domain-specific contexts without fine-tuning, as noted by Zhang et al. (2020).

### 2.4 Privacy-Preserving LLM Architectures

Emerging research has explored integrating LLMs with privacy-preserving techniques. Approaches such as retrieval-augmented generation (RAG) enable models to fetch domain-specific knowledge from external repositories without training on sensitive data (Lewis et al., 2020). Complementary mechanisms like Named Entity Recognition (NER) can anonymize input queries to safeguard PII (Li et al., 2021). Despite these innovations, a comprehensive framework combining privacy, regulatory compliance, and operational scalability for customer support remains underexplored.

## 2.5 Gaps and Contributions

While significant progress has been made in individual areas, there is limited work on unifying privacy-preserving techniques with zero-shot LLM capabilities in customer support. This publication addresses these gaps by proposing a framework that leverages zero-shot learning, real-time anonymization, and retrieval-augmented techniques to deliver privacy-compliant, scalable, and efficient customer support.

## 3. Privacy Challenges in Traditional Approaches

Machine learning (ML) systems traditionally rely on extensive training data to achieve high levels of accuracy and task-specific performance. In customer support, this often involves processing sensitive information, such as personally identifiable information (PII), financial details, or contractual agreements. While this data is critical for building and fine-tuning models, it also poses significant privacy risks. Mishandling of sensitive data can result in breaches, unauthorized access, and non-compliance with stringent regulations, such as the General Data Protection Regulation (GDPR) and the California Consumer Privacy Act (CCPA) (European Union, 2016; CCPA, 2018). Consequently, organizations face mounting pressure to adopt privacy-preserving practices while maintaining the quality of their ML applications.

One commonly used privacy-preserving technique is data anonymization, which involves removing or obfuscating identifiable elements in datasets. While this approach can reduce exposure risks, studies have shown that anonymized data can often be re-identified through auxiliary information or advanced statistical techniques, compromising its effectiveness (Rocher et al., 2019). Additionally, data anonymization frequently reduces the utility of the dataset, impacting the performance of ML models and making it unsuitable for complex tasks like customer support.

**Fig 3.1: Traditional Machine Learning Solution Development Approach**

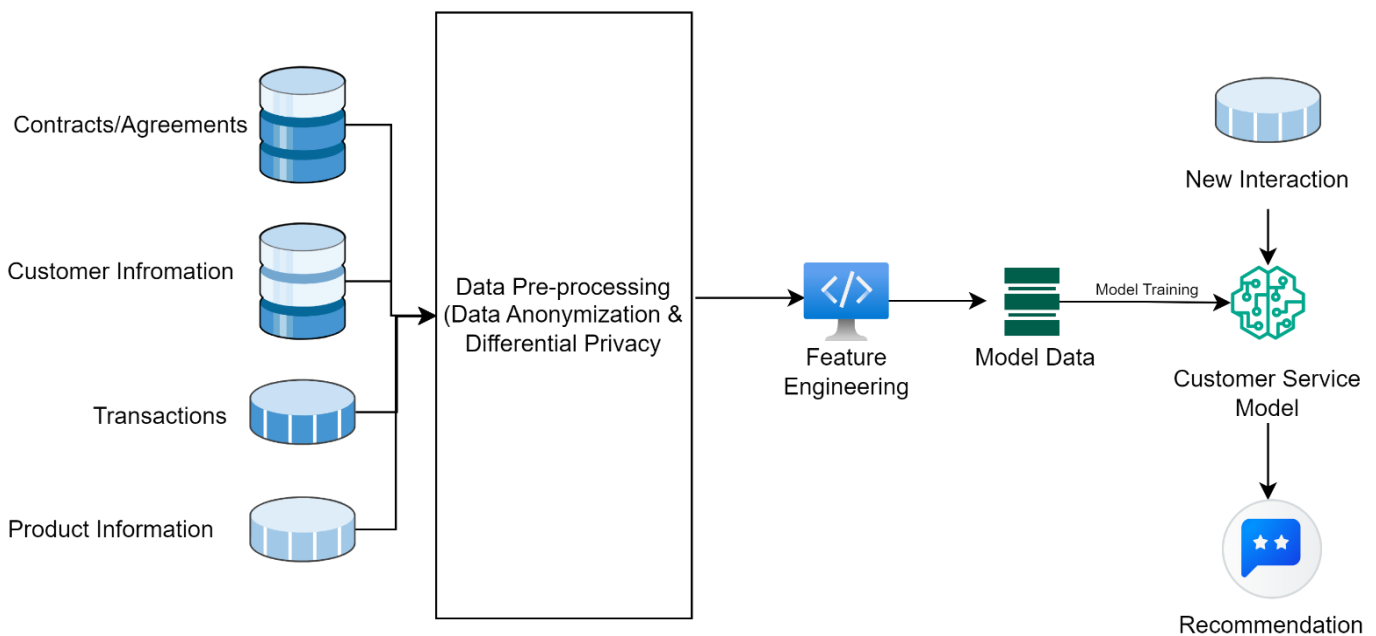

Differential privacy offers stronger theoretical guarantees by adding noise to data or model outputs, ensuring that individual contributions remain indistinguishable (Abadi et al., 2016). However, achieving an optimal balance between privacy and utility remains challenging. High levels of noise degrade model performance, which can lead to inaccurate or unhelpful customer support responses. Furthermore, differential privacy often requires significant computational resources, increasing the complexity and cost of implementing these techniques at scale.

Federated learning has emerged as a promising alternative to centralized training. It enables organizations to train models locally on user devices without transferring sensitive data to a central server (Kairouz et al., 2021). While this approach reduces the risks of data leakage during transmission, it introduces new challenges, such as communication overhead, model synchronization, and vulnerabilities in local devices. Moreover, federated learning does not inherently prevent sensitive information from being inadvertently encoded in the model, leaving potential exposure risks.

These limitations underscore the need for innovative solutions that minimize reliance on sensitive data while ensuring compliance with privacy regulations. Traditional approaches are either too resource-intensive, compromise data utility, or fail to provide comprehensive privacy guarantees. The present work addresses these challenges by introducing a zero-shot learning (ZSL) framework that leverages large language models (LLMs) without requiring local training on sensitive data, thereby reducing privacy risks and regulatory complexities.

## 4. Zero-Shot Learning (ZSL) as a Paradigm Shift

Zero-Shot Learning (ZSL) represents a transformative approach in natural language processing (NLP), enabling models to perform tasks without requiring additional task-specific training. This capability is particularly advantageous in scenarios where data collection and annotation are infeasible, costly, or pose significant privacy risks. Recent advancements in large language models (LLMs) like GPT-3 and T5 have demonstrated the potential of ZSL to generalize across diverse domains using knowledge acquired during pre-training on extensive datasets (Brown et al., 2020; Raffel et al., 2020). For customer support applications, this shift eliminates the traditional dependency on training models with organization-specific, potentially sensitive data, offering a significant leap in both operational efficiency and data privacy.

Unlike traditional machine learning (ML) methods, which require fine-tuning on domain-specific datasets, ZSL leverages the generalization ability of pre-trained models to handle a wide range of tasks. This approach aligns particularly well with customer support scenarios, where models must address varied user queries spanning multiple domains and knowledge areas. ZSL models achieve this by interpreting task instructions and contextualizing them within their pre-trained knowledge base, often formulated as a natural language prompt. For instance, a query about financial policies can be answered without the model ever having been trained on the organization's proprietary data, significantly reducing the risks associated with data leakage or breaches (Brown et al., 2020).

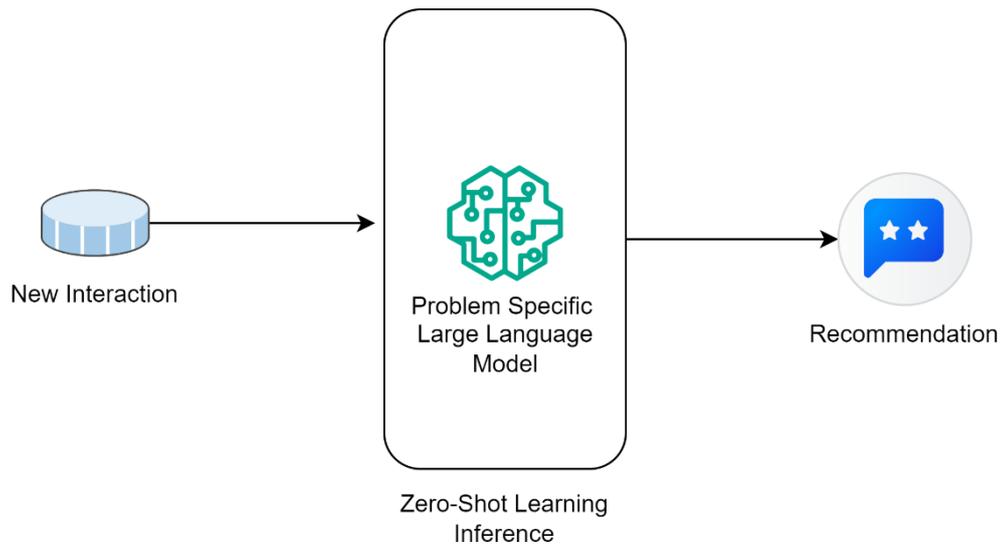

**Fig 4.1-- Zero Shot learning (Inference Process) Representation**

An additional strength of ZSL is its ability to reduce compliance complexities under privacy regulations such as the General Data Protection Regulation (GDPR) and the California Consumer Privacy Act (CCPA). Traditional ML models often necessitate storing and processing sensitive data locally, leading to challenges in fulfilling requirements such as the "right to be forgotten" or data minimization. By contrast, ZSL-based systems do not require sensitive data for training or storage, inherently addressing these regulatory concerns (European Union, 2016; CCPA, 2018).

However, ZSL does present challenges, particularly in domain-specific applications. While pre-trained models are proficient at general tasks, their performance may degrade when tasked with highly specialized queries or niche knowledge areas. To address these gaps, hybrid approaches, such as Retrieval-Augmented Generation (RAG), have been proposed. These methods supplement ZSL capabilities by incorporating external domain-specific knowledge bases to enhance response accuracy without requiring additional training on sensitive data (Lewis et al., 2020).

The adoption of ZSL in customer support not only revolutionizes operational workflows but also sets a new benchmark for privacy preservation in AI-driven systems. By removing the need for local training and integrating privacy-aware mechanisms like dynamic redaction and external knowledge retrieval, ZSL establishes itself as a key enabler of scalable, secure, and compliant AI solutions.

## 5. Proposed Framework: Privacy-Preserving Zero-Shot Learning (PP-ZSL)

The **Privacy-Preserving Zero-Shot Learning (PP-ZSL)** framework is designed to deliver accurate, scalable, and privacy-compliant responses in customer support. It ensures sensitive information is protected throughout the interaction pipeline. Below is a detailed discussion of each component of the framework, accompanied by the flowchart below.

Fig 5.1 Proposed Privacy-Preserving Zero-shot Learning Framework

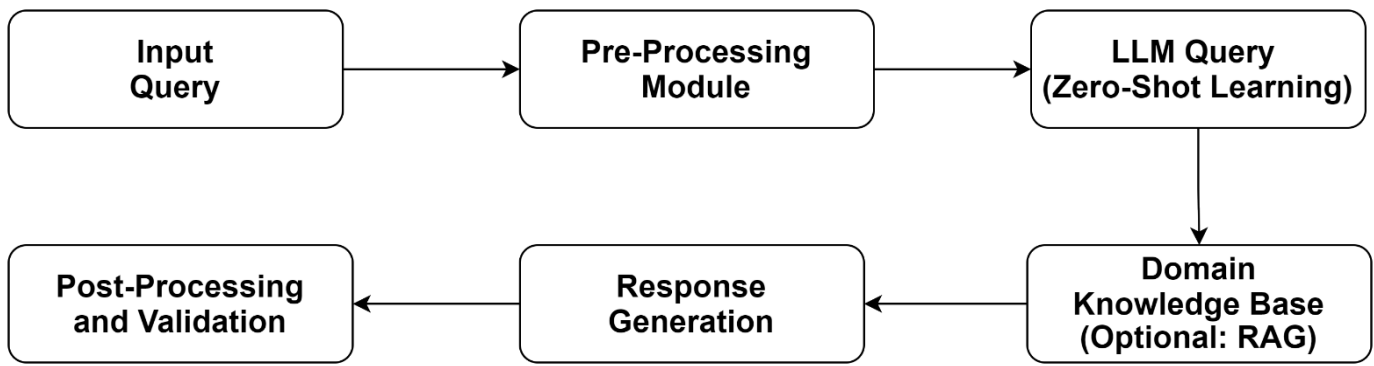

### 5.1 Input Query

The framework begins when a customer submits a query. This input often contains sensitive information, such as personally identifiable information (PII), financial data, or contractual details, which must be protected from the outset. The framework is designed to safeguard this data throughout the interaction pipeline, ensuring it is not exposed to unauthorized access. This initial step sets the stage for privacy-aware processing, aligning with stringent privacy regulations like GDPR and CCPA.

### 5.2 Pre-Processing Module

The pre-processing module serves as the first layer of privacy protection, ensuring that sensitive data in the input query is detected and anonymized before reaching the language model. Techniques such as Named Entity Recognition (NER) are employed to identify sensitive entities like names, account numbers, and dates. These entities are then either masked or replaced with placeholders through tokenization and redaction. The module also incorporates dynamic anonymization, which adjusts the level of redaction based on specific privacy policies or organizational requirements. By processing the query at this stage, the framework ensures that no raw sensitive data is passed to the LLM, reducing risks of inadvertent data exposure.

### 5.3 LLM Query (Zero-Shot Learning)

At the heart of the framework is the use of pre-trained Large Language Models (LLMs) operating in a zero-shot learning (ZSL) mode. Unlike traditional machine learning methods that require fine-tuning on organization-specific data, ZSL leverages the generalization capabilities of pre-trained models to interpret and respond to queries. This approach eliminates the need for local training, significantly reducing privacy risks and operational costs. The LLM uses its vast pre-trained knowledge to address a wide range of customer inquiries, ensuring accurate and contextually relevant responses without requiring sensitive data for additional training.

### 5.4 Domain Knowledge Base (Optional: RAG)

To enhance the domain-specific accuracy of responses, the framework optionally integrates a **retrieval-augmented generation (RAG)** module. This component augments the LLM's zero-shot capabilities by fetching relevant non-sensitive information from secure knowledge repositories. For example, queries about organization-specific policies or technical details can be addressed by retrieving precise, up-to-date information from external databases or APIs. This integration ensures that the LLM remains contextually accurate while avoiding the need to store or process sensitive domain-specific data locally.

### 5.5 Response Generation

Once the query is processed, the LLM generates a response based on the anonymized input and any supplementary information from the RAG module. The generated response is designed to be contextually appropriate and tailored to the customer's query. By preserving the anonymization applied during pre-processing, the response avoids reintroducing sensitive data, maintaining privacy throughout the interaction.

### 5.6 Post-Processing and Validation

The final step involves a robust post-processing and validation module, which ensures that the generated response adheres to privacy and regulatory standards. **Privacy filters** are applied to detect and remove any inadvertently reintroduced sensitive data. Additionally, a **compliance audit** is conducted to validate that the response aligns with organizational and legal policies, such as GDPR's "right to be forgotten" or CCPA's data minimization requirements. This step guarantees that the final response is both privacy-compliant and operationally effective, ready for delivery to the customer.

The PP-ZSL framework integrates these components into a seamless workflow, addressing the dual challenges of data privacy and operational scalability. By anonymizing queries, leveraging zero-shot learning, and ensuring rigorous validation, the framework offers a transformative solution for AI-driven customer support that balances accuracy, efficiency, and compliance.

The framework ensures end-to-end privacy preservation while delivering accurate and efficient responses. It begins with input query anonymization, processes the query using pre-trained LLMs with optional domain-specific augmentation, and concludes with post-processing validation to ensure compliance.

This modular approach allows organizations to implement the framework flexibly and adapt it to specific operational or regulatory needs.

# 6. Future Directions

The proposed Privacy-Preserving Zero-Shot Learning (PP-ZSL) framework provides a foundation for addressing privacy challenges in customer support, but several avenues for future research and development can enhance its scalability, adaptability, and effectiveness. This section discusses key directions, focusing on hybrid approaches, real-time privacy enhancements, contextual understanding, and compliance with emerging global privacy laws.

## 6.1 Hybrid Approaches for Domain-Specific Adaptability

While zero-shot learning (ZSL) enables generalization across domains, its performance may degrade in highly specialized contexts. Future research could explore **hybrid models** that combine ZSL with lightweight fine-tuning on anonymized or synthetic data. Techniques such as prompt engineering (Brown et al., 2020) or instruction-tuning (Wei et al., 2022) can help bridge the gap between generalization and domain specificity. Synthetic datasets generated using privacy-preserving methods like differential privacy (Abadi et al., 2016) or generative adversarial networks (GANs) (Goodfellow et al., 2014) could provide a safe medium for task-specific training without compromising sensitive information.

## 6.2 Advances in Real-Time Privacy Filters

Improving real-time mechanisms for data anonymization and redaction is critical for practical deployment. Tools like Named Entity Recognition (NER) models can be enhanced to detect and anonymize sensitive data with greater accuracy and context-awareness (Li et al., 2021). Future work could also explore multi-modal privacy filters, integrating textual, visual, and auditory data anonymization to support diverse customer support channels. Leveraging on-device processing for privacy filtering could further reduce latency and enhance data security.

## 6.3 Enhancing Contextual Understanding in LLMs

One of the limitations of current LLMs is their struggle with multi-turn interactions and maintaining contextual understanding in conversations. Future research could integrate **context-tracking mechanisms** that allow models to retain and utilize dialogue history securely. Techniques such as retrieval-augmented generation (RAG) (Lewis et al., 2020) could enable models to fetch relevant contextual information dynamically, reducing reliance on sensitive local datasets.

## 6.4 Addressing Scalability and Cost-Efficiency

Deploying ZSL frameworks at scale in real-time customer support systems remains a challenge due to computational demands. Future work could explore the integration of **model optimization techniques** such as quantization (Gong et al., 2014) and pruning (Han et al., 2015) to reduce inference costs without sacrificing performance. Additionally, designing serverless architectures that offload model execution to edge devices could balance scalability and cost.

### 6.5 Adapting to Emerging Privacy Regulations

As global privacy laws evolve, it is crucial to ensure that the PP-ZSL framework remains compliant with new regulations. Future research could focus on creating **dynamic compliance modules** that adapt to region-specific privacy requirements in real time. For example, laws like India's Data Protection Act and China's Personal Information Protection Law require localized adaptations that balance compliance and operational efficiency (Chik, 2021). A key direction is the integration of automated compliance monitoring systems that flag potential violations in model outputs.

### 6.6 Ethical Considerations and Bias Mitigation

LLMs often exhibit biases inherited from pre-training datasets, which can lead to discriminatory or unethical outputs (Binns et al., 2018). Future research could explore methods for **bias detection and mitigation** that align with privacy-preserving principles. Techniques like adversarial training (Zhang et al., 2018) or counterfactual data augmentation (Kaushik et al., 2020) could be adapted to improve model fairness without exposing sensitive information.

### 6.7 Explainability and Trust in AI

Building trust in AI systems requires transparency and explainability. Future work could focus on developing **interpretable AI mechanisms** tailored to privacy-preserving frameworks. For instance, techniques like SHAP (Lundberg & Lee, 2017) or LIME (Ribeiro et al., 2016) could be extended to explain LLM decisions in real-time customer support interactions while ensuring privacy.

### 6.8 Exploring Cross-Domain Applications

While this framework focuses on customer support, its principles can be extended to other domains, such as healthcare, finance, and legal services. Each domain presents unique privacy challenges and regulatory requirements, making cross-domain adaptability a rich area for exploration. For example, applying ZSL in healthcare could leverage privacy-preserving techniques to anonymize patient data while enabling accurate diagnosis and treatment recommendations (Rieke et al., 2020).

## 7. Conclusion:

The **Privacy-Preserving Zero-Shot Learning (PP-ZSL) framework** is a game-changer for AI-ready organizations, offering a scalable, efficient, and privacy-compliant approach to customer support. By eliminating the need for local training on sensitive datasets, the framework significantly minimizes data privacy risks while ensuring compliance with regulations like GDPR and CCPA. Its integration of real-time anonymization safeguards sensitive information, and zero-shot learning capabilities allow pre-trained LLMs to deliver accurate, contextual responses without additional fine-tuning. This not only enhances customer satisfaction but also reduces operational overhead, enabling rapid deployment and significant cost savings.

The framework's dynamic knowledge integration through retrieval-augmented generation (RAG) ensures adaptability across domains, making it suitable for organizations operating in diverse industries. It simplifies compliance reporting by aligning with privacy-by-design principles, reducing legal risks, and future-proofing operations against evolving regulatory landscapes. By focusing on modularity and innovation, the PP-ZSL framework empowers businesses to allocate resources toward strategic objectives while offering consistent, high-quality customer support across channels.

In essence, PP-ZSL equips AI-ready organizations to meet the dual demands of privacy and performance. Whether addressing sensitive financial queries, healthcare support, or e-commerce assistance, the framework balances operational efficiency, customer experience, and data security, positioning businesses to thrive in a privacy-conscious digital landscape.

## 8. Potential Use Cases:

The **Privacy-Preserving Zero-Shot Learning (PP-ZSL) framework** has broad applicability across diverse industries where customer support demands high accuracy, scalability, and strict adherence to privacy regulations. In the **financial services** sector, the framework can securely handle banking and insurance queries, such as account details, loan applications, and fraud detection support. It ensures that sensitive financial data and personal identifiers are protected while providing real-time, context-aware responses to customer inquiries.

In **healthcare**, the framework can assist patients with general medical advice, appointment scheduling, and test result clarifications without accessing sensitive health records. For telemedicine services, it enables virtual consultations while preserving patient privacy and confidentiality. Similarly, in **e-commerce**, the PP-ZSL framework can manage inquiries related to orders, returns, and personalized product recommendations using anonymized customer preferences, ensuring a seamless and secure shopping experience.

For **legal and contractual support**, the framework offers the ability to analyze contracts, address questions about legal clauses, and resolve disputes without exposing sensitive legal or contractual data. **Telecommunication providers** can leverage it to assist customers with technical issues, billing inquiries, and service outages, all while protecting device identifiers and account details. **Retail and loyalty programs** can benefit from the framework by securely managing loyalty rewards, promotional offers, and customer feedback while safeguarding identifiable information.

In **government and public services**, the framework can support citizen inquiries about services, benefits, or taxes while maintaining the confidentiality of personal and financial data. It also facilitates regulatory compliance assistance, helping citizens understand policy changes securely. For **education and e-learning**, the framework can address student inquiries about enrollment, courses, and grades, and offer learning assistance without compromising personal data.

The PP-ZSL framework is equally valuable in **human resources** for responding to employee questions about payroll, benefits, and leave policies while ensuring data privacy. It can also support recruitment processes by providing anonymized feedback on applications and interviews. In the **travel and hospitality** sector, the framework can assist with travel bookings, cancellations, and itinerary changes while protecting sensitive payment and travel details, enabling personalized travel recommendations securely.

For **real estate**, the framework facilitates secure property-related queries, pricing information, and mortgage assistance, protecting sensitive financial and personal details. It also excels in **customer feedback and analytics**, enabling secure aggregation, analysis, and sentiment measurement of customer feedback while ensuring data anonymity.

By ensuring compliance with privacy laws like GDPR and CCPA, the PP-ZSL framework offers a secure, scalable, and efficient solution for customer support across industries. It reduces the risks associated with handling sensitive data, enhances the customer experience through secure, real-time interactions, and serves as a versatile tool for any domain requiring privacy-focused support.

## 9. Conflict of Interest:

Authors do not have any conflict of interest as there is no external/internal funding used to complete this work.